\documentclass{article}

    \usepackage[preprint,nonatbib]{neurips_2020}

\usepackage[utf8]{inputenc} 
\usepackage[T1]{fontenc}    
\usepackage{hyperref}       
\usepackage{url}            
\usepackage{booktabs}       
\usepackage{amsfonts}       
\usepackage{nicefrac}       
\usepackage{microtype}      

\usepackage{amsmath}
\RequirePackage{amssymb}
\RequirePackage{amsbsy}
\usepackage{algpseudocode}
\usepackage{algorithm}
\usepackage{subcaption}
\usepackage[usenames,dvipsnames]{xcolor}
\usepackage{enumitem}
\usepackage[title]{appendix}
\usepackage{graphics}
\usepackage{graphicx}
\usepackage{tikz}
\usetikzlibrary{arrows.meta}
\usetikzlibrary{decorations.pathreplacing}

\renewcommand{\vec}[1]{\ensuremath{\boldsymbol{#1}}}
\newcommand{\vechat}[1]{\ensuremath{\vec{\hat #1}}}
\newcommand{\CM}{\ensuremath{\mathbb{F}}}

\newcommand{\angles}[1]{\left< #1 \right>}

\newcommand{\braces}[1]{\left\{ #1 \right\}}

\providecommand{\abs}[1]{\lvert#1\rvert}
\providecommand{\norm}[1]{\lVert#1\rVert}

\title{A Linear Algebraic Approach \\to Model Parallelism in Deep Learning}

\author{%
  Russell J.~Hewett\thanks{https://rjh.io} \\
  Department of Mathematics\\
  Virginia Tech\\
  Blacksburg, VA, 24061 \\
  \texttt{rhewett@vt.edu} \\
  \And
  Thomas J. Grady II \\
  \texttt{thomasg2@vt.edu} \\
}

\begin{document}

\maketitle

\begin{abstract}
  Training deep neural networks (DNNs) in large-cluster computing environments is increasingly necessary, as networks grow in size and complexity.  Local memory and processing limitations require robust data and model parallelism for crossing compute node boundaries. We propose a linear-algebraic approach to model parallelism in deep learning, which allows parallel distribution of any tensor in the DNN.
  Rather than rely on automatic differentiation tools, which do not universally support distributed memory parallelism models, we show that parallel data movement operations, e.g., broadcast, sum-reduce, and halo exchange, are linear operators, and by defining the relevant spaces and inner products, we manually develop the adjoint, or backward, operators required for gradient-based training of DNNs.
  We build distributed DNN layers using these parallel primitives, composed with sequential layer implementations, and demonstrate their application by building and training a distributed DNN using DistDL, a PyTorch and MPI-based distributed deep learning toolkit.
\end{abstract}

\section{Motivation \& Background}

Training deep neural networks (DNNs) on extreme-scale super computers is a challenging problem, however, it is increasingly a necessary component of modern computational and data science workflows. For extremely large problems in scientific machine learning (SciML; e.g., those in physics-guided ML~\cite{KarpatneAtluriEtAl2017} which require integration of parallel partial differential equation (PDE) solvers)~\cite{KurthTreichlerEtAl2018a,YangTreichlerEtAl2019b} and other large multi-dimensional or volumetric data processing problems, such as video processing or seismic data processing, limitations on local memory and processing power (even with modern large memory GPUs) require robust parallelism models to cross the compute-node boundary. Data parallelism is ubiquitous in deep learning, but model parallelism has been harder to achieve.  In particular, this is because the ``model'' in large deep neural networks is highly irregular and has no uniform spatial structure to induce the sparseness that is typical in large-scale parallel problems.  To achieve a fully parallelized deep neural network training algorithm, we focus on enabling parallelism by distributing any tensor in a network including learnable parameters, inputs, and outputs.

Recently, multiple frameworks have been developed which partially address the distributed deep learning problem.  These frameworks build from, or into, popular frameworks such as PyTorch~\cite{Ketkar2017} and Tensorflow~\cite{AbadiBarhamEtAl2016}, which natively support data parallelism, to add support for pipelining~\cite{Microsoft2020}, or limited support for some model parallelism over some aspects of the network~\cite{ShoeybiPatwaryEtAl2019,SergeevDelBalso2018,Shazeer2019,ShazeerChengEtAl2018}.  Native support for distributed learning is also slowly appearing in the popular frameworks.   Parallelism in individual aspects of deep learning, such as convolutional layers has also been investigated~\cite{DrydenMaruyamaEtAl2019a}, as well as application of some applications to parallel physics-driven network structures in, e.g., seismic inversion~\cite{Richardson2018}.  However, current approaches only provide partial solutions and we lack a complete, integrated framework for treating the distributed learning problem.  Here, we aim to provide such a framework.  While this manuscript generally addresses ``model'' parallelism within a single network gradient calculation, our framework readily admits classical data parallelism and pipelining.

Automatic (or algorithmic) differentiation (AD) is among the most important tools that computational science has contributed to the democratization of deep learning.  Given an implementation of a computer algorithm for evaluating a non-linear function $\mathcal{F}$, forward-mode AD produces an implementation of an algorithm for evaluating the action of the $F$, the Jacobian of $\mathcal{F}$, and backward- or adjoint-mode AD produces an algorithm for evaluating the action of $F^*$, the adjoint of the Jacobian of $\mathcal{F}$~\cite{Naumann2012}.  In computational science problems, such as PDE-constrained optimization~\cite{Plessix2006}, AD is frequently applied to forward computation kernels to develop correct adjoint kernels necessary for numerical optimization.  In deep learning, it is used for similar tasks in the construction of gradient calculations needed to invert for the parameters in composite non-linear functions, such as DNNs.

However, AD tools, especially those in widely used deep learning frameworks, have limited support for the message-passing operations required to run codes on distributed memory supercomputers, hampering the development of fully parallel deep learning codes.  Historically, some AD tools~\cite{BischofRohEtAl1997} have provided limited support for differentiating distributed memory parallel codes, e.g., enabled via the Message Passing Interface (MPI)~\cite{ClarkeGlendinningEtAl1994}.  However, such support is not ubiquitous.  Fortunately, as we will show, the operations necessary for distributed memory parallelism are linear.  Consequently, we do not need to appeal to AD to generate the adjoint operations needed for gradient calculation.  Instead, we exploit the definition of the adjoint operator, careful definitions of the spaces they act upon, and the inner products on those spaces, to build a set of primitive operations, and their adjoints, to describe {data movement}\footnote{We avoid the term \emph{communication} because our model applies beyond classical distributed memory settings.}
in computers and distributed memory supercomputers.

As we will demonstrate, these operations can be embedded into a deep learning framework using the framework's native interface for specifying new functions, and composited with existing network layers or functions.  Thus, the data movement operations, and their adjoints, become merely another function for the framework's automatic differentiation tool to operate on.  We have implemented a proof-of-concept in our distributed deep learning tool, DistDL, using MPI (via mpi4py~\cite{DalcinPazEtAl2011}) and PyTorch for CPU functions.  This restriction is not a limitation of our model, only a restriction of our current implementation: there are no major technological impediments to transitioning this model to large-scale hybrid CPU-GPU supercomputers.


\section{Linear algebraic memory model}\label{sec:memory_model}

Let $\CM$ be the space of relevant computer numbers, e.g., integers or IEEE floating point numbers.  If $\mathcal{F}: \CM^m \rightarrow \CM^n$ is a linear operator, then $F = \mathcal{F}$ is its Jacobian and the adjoint of the Jacobian, $F^*$, is defined by the adjoint relationship,
\begin{align}
  \angles{F\vec{x}, \vec{y}}_{\CM^n} = \angles{\vec{x}, F^*\vec{y}}_{\CM^m},
\end{align}
where $\CM^k$ represents a $k$-length subset of a computer's memory.  For the purposes of this development, we take the inner product to be the standard Euclidean inner product,\footnote{When $\CM$ is the space of floating point numbers, the inner product must be constructed carefully, especially in parallel environments, because floating point arithmetic is not commutative.}
\begin{align}
\angles{\vec{a}, \vec{b}}_{\CM^k} = \sum_{i=0}^{k-1}a_ib_i\;\;\vec{a},\,\vec{b}\in\CM^k.
\end{align}
Thus, with a concrete implementation of $F$, we can derive and implement concretely the coherent associated $F^*$, and we can exploit these implementations in a deep learning framework's AD tool, such as PyTorch's \texttt{autograd}.  In general, the data realized in the subsets of the memory are subsets of tensors.  In defining these operations, we make no assumptions about the rank, ordering, size, or layout of the tensor, though these do matter in a practical implementation.

To build parallel primitives for deep learning, we must first understand the nature of $\CM^k$ and of the operators on it.  In the ensuing discussion, we consider the concept of ``a computer's memory'' to be extremely inclusive.  While it is easiest to consider $\CM^k$ to be the main memory of a single CPU of a single compute node (or worker), this framework admits auxiliary memories, such as those attached to GPU accelerators, remote memory on other compute nodes or cloud instances, or even disk.

Here, we develop linear representations of primitive memory operations and their adjoints, which we will use to develop parallel data movement primitives and more complex distributed neural network layer structures.  We are careful to point out that the manual procedure that we outline is essentially how adjoint-mode AD works.  However, we find it useful to view these operations from a linear-algebraic perspective, rather than from the typical computation-graph perspective used in AD.  Most AD tools are generally incapable of handling \emph{all} possible data movement operations within our inclusive memory model, so we must be able to build the operations manually.  Thus, our framework provides the theoretical glue necessary to implement these operations when they are not available natively.  Moreover, in manual implementations we can make some optimizations that AD-generated codes cannot make, as some operations appear only implicitly in forward codes, but must appear explicitly in adjoint codes, or vice versa.  In this manuscript we err on the side of being explicit, while practical implementations may not explicitly include all operations, except during validation.

\paragraph{Allocation}

The \textit{allocation} of a subset of memory, to be realized by $\vec{x}_b = \vec{0}_b$, for a program that already has space for $\vec{x}_a$ available, is a linear operation $A_{b}: \CM^m \rightarrow \CM^n$,
\begin{align}
  A_{b}\vec{x} = \begin{bmatrix}I_{a} \\ O_{b} \end{bmatrix}\begin{bmatrix}\vec{x}_{a} \end{bmatrix} = \begin{bmatrix}\vec{x}_{a} \\ \vec{0}_{b} \end{bmatrix},
\end{align}
where $I_{a}$ is an identity operator on the original subset and $O_{b}$ is a zero operator on the new subset.  The adjoint of allocation, $A^*_b$, is derived through the standard inner product, which we detail in Appendix~\ref{app:derivations}.  $A^*_b$ is the transpose\footnote{The adjoint is strongly dependent on the inner product and is not always the matrix transpose.} of $A_b$, and acts on a realization $\vec{y}$ from $\CM^n$,
\begin{align}
  A^*_b \vec{y} = A_b^T\vec{y} = \begin{bmatrix}I_{a} & O_{b} \end{bmatrix}\begin{bmatrix}\vec{y}_{a} \\ \vec{y}_{b} \end{bmatrix} = \begin{bmatrix}\vec{y}_{a} \end{bmatrix} .
\end{align}
The adjoint of allocation is thus \textit{deallocation}, and similarly the deallocation primitive $D_b$ has allocation as its adjoint, $D_b^* = A_b$.

We use a liberal definition of ``allocation'' that goes beyond classical memory allocation operations (e.g., \texttt{malloc()} in C) because these operations are describing the semantics of an implementation, not syntax.  Allocation is any operation which brings memory into scope, including formal allocation, the addition of data to the stack, creation of a reference, etc.  In the context of a neural network layer, this means that if data is not checkpointed for use in the adjoint phase during the forward phase, it goes out of scope or is ``deallocated'' when the forward function completes.

\paragraph{Clear}
The \textit{clear} operator, $K_b$, sets a realization of a subset of $\vec{x}$, $\vec{x}_b$ to \vec{0}.  The operation, $K_b:~\CM^m~\rightarrow~\CM^m$ is realized by,
\begin{align}
  K_b\vec{x} = \begin{bmatrix}I_a & \\ & O_b \end{bmatrix}\begin{bmatrix}\vec{x}_a \\ \vec{x}_b\end{bmatrix} = \begin{bmatrix}\vec{x}_a \\ \vec{0}_b\end{bmatrix},
\end{align}
and it is trivially self-adjoint, $K_b^* = K_b$.

\paragraph{Add}
The \textit{add} operator, $S_{a \rightarrow b}: \CM^{m} \rightarrow \CM^{m}$, performs in-place summation $\vec{x}_a$ to $\vec{x}_b$,
\begin{align}
  S_{a\rightarrow b}\vec{x} = \begin{bmatrix} I_a & \\ I_a & I_b \end{bmatrix}\begin{bmatrix}\vec{x}_a \\ \vec{x}_b\end{bmatrix} = \begin{bmatrix}\vec{x}_a \\ \vec{x}_a + \vec{x}_b\end{bmatrix}.
\end{align}
The adjoint of an add is also an add, but in the reverse direction,
\begin{align}
  S^*_{a\rightarrow b}\vec{y} = \begin{bmatrix} I_a & I_b \\ & I_b \end{bmatrix}\begin{bmatrix}\vec{y}_a \\ \vec{y}_b\end{bmatrix} = \begin{bmatrix}\vec{y}_a + \vec{y}_b \\ \vec{y}_b\end{bmatrix} = S_{b\rightarrow a}\vec{y}.
\end{align}

\paragraph{Copy}

The \textit{copy} operator, which copies data from the subset $\vec{x}_a$ to $\vec{x}_b$, has both in-place and out-of-place forms, made distinct only by the semantics of an implementation.  An in-place copy is the composition of clear and add while an out-of-place copy is the composition of allocate and add.  This may seem pedantic, as ``\texttt{x = c;}'' is more concise than ``\texttt{x = x * 0; x = x + c;}'', but the distinction becomes important for defining higher-level operations.  We define the operators and adjoints below, and justify the construction in Appendix~\ref{app:derivations}.
\begin{align*}
  &\text{In-place Copy}                           & &\text{Out-of-place Copy} \\
  &C_{a\rightarrow b}   = S_{a\rightarrow b}K_b  & &C_{a\rightarrow b} = S_{a\rightarrow b}A_b\\
  &C^*_{a\rightarrow b} = K_bS_{b\rightarrow a}  & &C^*_{a\rightarrow b} = D_bS_{b\rightarrow a}
\end{align*}
The choice of in-place or out-of-place operation impacts only implementation decisions when defining higher-level operations, so we do not distinguish them in the sequel.

\paragraph{Move}

The \textit{move} operator moves a realization $\vec{x}_a$ to $\vec{x}_b$, and similar to copy, has in-place and out-of-place forms given below, which we justify in the  Appendix~\ref{app:derivations}.
\begin{align*}
  &\text{In-place Move}                           & &\text{Out-of-place Move} \\
  &M_{a\rightarrow b}   = K_aS_{a\rightarrow b}K_b  & &M_{a\rightarrow b} = D_aS_{a\rightarrow b}A_b\\
  &M^*_{a\rightarrow b} = K_bS_{b\rightarrow a}K_a = M_{b \rightarrow a}  & &M^*_{a\rightarrow b} = D_bS_{b\rightarrow a}A_a = M_{b \rightarrow a}
\end{align*}
Again, the choice of in-place or out-of-place forms impacts only some implementation decisions.

\section{Linear algebraic primitives for data movement}\label{sec:parallel_primitives}

Using these primitive memory operations, we construct linear operators representing several standard parallel data movement primitives and their adjoints.  To accommodate operations on distributed memory computers, we now consider the definition of the memory space to include all memories on a compute cluster.  While we discuss operations as if the parallel workers are distinct compute nodes, this distinction is made explicit only by the communication library, such as MPI, and our model is independent of communication back-end.  In the ensuing discussion, we will generally assume that operations are out-of-place -- communicating data results in a new memory allocation on the ``receiving'' worker.  While this is generally not best practice in large-scale simulation, out-of-place operations better fit PyTorch's computation and AD model.
The one exception in this presentation is the halo exchange, which we will describe as an in-place operation, following from standard practice in large-scale simulation.  Adapting internal mechanics of out-of-place operations to in-place operations has no bearing on the outcome.  Thus, if such an implementation is preferred in performance environments, it is of minor consequence.

For brevity, we will not show allocations or deallocations, but their implicit presence may be felt.  While we generally express data movement using copy, if the primal realization is deallocated without further use after the copy, the copy may be expressed as a move.  In practice, many operations we make explicit are needed only theoretically.  For example, in the adjoint halo exchange we express clears on the exchange buffers for mathematical consistency, but these are handled implicitly when assigning data to the buffers.


\paragraph{Send and Receive}

The most basic distributed memory data movement operation, from which all others can be derived, is the \textit{send-receive} operator.  In a concrete implementation, the send-receive pair requires two function calls (send and receive) by separate workers, but from a linear-algebraic perspective, the send-receive operator is simply a copy $C_{a\rightarrow b}$, where the subsets $\vec{x}_a$ and $\vec{x}_b$ are on the two different workers.  Consequently, the adjoint also follows from above.
While the send-receive operation is not self-adjoint, a practical implementation of its adjoint requires a receive-send pair, but the add operation may not be equivalent to assignment, as it is in the forward operation.

\paragraph{Scatter and Gather}

The \textit{scatter} primitive is essentially a sequence of send-receive pairs, where subsets of $\vec{x}_a$ are copied to multiple other workers.  Linear-algebraically, this is is a block-diagonal matrix with send-receive blocks.  The adjoint derivation follows from the previous discussion.  If the data movement operations are equivalent to move, then the adjoint operation becomes an instance of the \textit{gather} primitive, which collects data from multiple workers into one subset on one worker, otherwise communication still follows the gather pattern but the summation must be respected.

\paragraph{Broadcast}

A critical parallel primitive, the \textit{broadcast}, is identified and implemented in many distributed memory deep learning tools~\cite{Ketkar2017,ShoeybiPatwaryEtAl2019,SergeevDelBalso2018} because it is necessary to distribute network parameters to multiple workers.  A broadcast, $B_{a \rightarrow \braces{k}}$, is a linear operator from a one realization on subset $\vec{x}_a$ to $k$ realizations on subsets $\vec{x}_0, \dots \vec{x}_{k-1}$ and is $k$ copy operations,
\begin{align}
  B_{a \rightarrow \braces{k}}\vec{x}_a = \begin{bmatrix}C_{a \rightarrow 0} \\ C_{a \rightarrow 1} \\ \vdots \\ C_{a \rightarrow k-1} \end{bmatrix} \vec{x}_a = \begin{bmatrix}\vec{x}_a \\ \vec{x}_a \\ \vdots \\\vec{x}_a \end{bmatrix} = \vec{x}_{\braces{k}}.
\end{align}
While the above ``implementation'' scales linearly with $k$, the canonical logarithmic broadcast implementation has an equivalent representation.
For in-place versions, the first copy is an identity operator.
Then, the adjoint of the broadcast is,
\begin{align}
  B^*_{a \rightarrow \braces{k}}\vec{y}_{\braces{k}} &= \begin{bmatrix}C^*_{a \rightarrow 0} & C^*_{a \rightarrow 1} & \cdots & C^*_{a \rightarrow k-1} \end{bmatrix}\vec{y}_{\braces{k}}
                                                     = \sum_{i=0}^{k-1} K_i S_{i \rightarrow a} \vec{y}_i
                                                     = \vec{y}_a.
\end{align}
The summation term is the key to understanding the adjoint broadcast implementation: the adjoint of the broadcast is a sum-reduction.

\paragraph{Sum-reduce and all-reduce}

The \textit{sum-reduce} primitive, of equal importance with the broadcast, represents summation of $k$ subsets into $\vec{x}_a$ and its derivation follows the reverse of the broadcast.  The sum-reduce operator, $R_{\braces{k}\rightarrow a} = B^*_{a \rightarrow \braces{k}}$, and its adjoint, $R^*_{\braces{k}\rightarrow a} = B_{a \rightarrow \braces{k}}$, is a broadcast.

While the \textit{all-reduce} operator is not necessary in our implementation, some distributed convolution formulations make use of the operation~\cite{DrydenMaruyamaEtAl2019a}.  In our framework, an all-reduce is simply the composition of a sum-reduce and a broadcast, $\mathcal{A}_{\braces{k}\rightarrow\braces{k}} = B_{a \rightarrow \braces{k}}R_{\braces{k} \rightarrow a}$, and is trivially self-adjoint, as $\mathcal{A}^*_{\braces{k}\rightarrow\braces{k}} = R^*_{\braces{k} \rightarrow a}B^*_{a \rightarrow \braces{k}} = B_{a \rightarrow \braces{k}}R_{\braces{k} \rightarrow a} = \mathcal{A}_{\braces{k}\rightarrow\braces{k}}$.

\paragraph{Generalized all-to-all}


In a DNN layer, the input, output, and parameter tensors have different dimension and shape, which strongly influences
load balance.  Consequently, parallel performance may require a change in a tensor's parallel decomposition when composing layers.  This is performed by an \textit{all-to-all} operation, which takes the appearance of a matrix transpose, and is also referred to as a shuffle~\cite{DrydenMaruyamaEtAl2019a}.
For generalized tensors with generalized partitions, data stored in one worker's memory may need to be copied to any other worker in the destination partition, essentially a scatter operation.  Then, in a linear algebraic sense, the all-to-all operation is a block permutation matrix, where the blocks are send-receive operators for all simultaneous scatters.  A similar result as for gathers above, holds for the adjoint of all-to-all.

\paragraph{Halo exchange}


In classical large-scale simulation, a decomposition of the relevant spatial domain allows for effective \textit{model parallelism}: large variables are distributed to different workers according to the spatial decomposition.  When a differential operator is sparse, physical interactions are local and minimal data, found near the domain boundaries, needs to be shared between adjacent workers.  In neural networks, analogous situations arise for layers featuring small, sliding kernels, such as convolutional layers and pooling layers. For each worker to correctly apply the computational kernel, this halo region must contain copies of the current data owned by neighboring workers.  The exchange of this boundary data between workers is known as a \textit{ghost exchange} or \textit{halo exchange}.

In finite difference-based simulation, the halo regions are regular in size.  In~\cite{DrydenMaruyamaEtAl2019a}, a halo exchange algorithm for convolutional kernels is presented in the context of a convolutional layer, assuming similar regularity.  Compact, centered kernels, such as convolutional kernels without striding or dilation, with carefully chosen domain decompositions will tend to have regularly sized halo regions.  However, for many use common use-cases in deep learning, e.g., for one-sided pooling kernels, for centered kernels without tailored partition sizes, and when load balance is driven by the \textit{output} tensor, we have observed that halo regions can have unbalanced structure.  We have illustrated a number of examples of this irregular structure in Appendix~\ref{app:halo}.


Due to this irregular structure, we use the linear-algebraic framework to define an algorithm for generalized halo exchange in distributed deep learning, as well as its adjoint.  In our algorithm, we neither make any assumptions on the rank of the input and output tensors (only that they are the same) nor the structure of the kernel.  As computational load on a given worker is driven by the volume of that worker's output subtensor,\footnote{This is also true for standard simulations, but the data sizes are generally fixed over a single time-step, so we do not usually think this way.} we assume that the output tensor is optimally load balanced and derive the necessary input tensor halo sizes in each dimension from there.  We assume that the tensors are sensibly decomposed, relative to kernel size, so that halos require data from directly adjacent neighbor workers only.  The portion of the distributed tensor that is owned by a worker is the bulk region and the halo exchange ensures that a worker has copies of the necessary portions of its neighbor's bulk regions in its halo region.

All halo regions, both left and right, from all dimensions of a rank-$d$ tensor may have different thickness.  The thicknesses are determined by the minimum and maximum global indices of the worker's output tensor and the size, stride, dilation, and padding parameters of the kernel.  From a linear-algebraic perspective, the halo exchange is a sequence of send-receive operations.  For efficiency, we assume that the halo exchange is in-place, so the input and output realizations are on the same memory subset.  Following the linear-algebraic view, the halo exchange operator for one worker exchanging with its neighboring workers in one dimension is,
  \begin{equation}
      H =
      K_{\textbf{T}}
      C_{\textbf{U}}
      C_{\textbf{E}}
      C_{\textbf{P}}
      K_{\textbf{S}},
  \end{equation}
  where
  $K_{\textbf{S}}$ the setup operator, clears the exchange buffers,
  $C_{\textbf{P}}$ the pack operator, copies from the bulk region to the send buffer,
  $C_{\textbf{E}}$ the exchange operator, copies from the current worker's send buffer to the neighboring worker's receive buffer, and vice versa,
  $C_{\textbf{U}}$ the unpack operator, copies from the receive buffer to the halo region,
  $K_{\textbf{T}}$ the teardown operator, clears on the exchange buffers.
For $d$-rank tensors, the exchange is performed one dimension at a time, in a nested manner to ensure proper communication of data in corner cases \cite{Gropp2016}.  Thus, the full exchange operator is,
  \begin{equation}
      H =
      K_{\textbf{T}_{d-1}}
      C_{\textbf{U}_{d-1}}
      C_{\textbf{E}_{d-1}}
      C_{\textbf{P}_{d-1}}
      K_{\textbf{S}_{d-1}}
      \dots
      K_{\textbf{T}_1}
      C_{\textbf{U}_1}
      C_{\textbf{E}_1}
      C_{\textbf{P}_1}
      K_{\textbf{S}_1}
      K_{\textbf{T}_0}
      C_{\textbf{U}_0}
      C_{\textbf{E}_0}
      C_{\textbf{P}_0}
      K_{\textbf{S}_0}
  \end{equation}
    with corresponding adjoint,
    \begin{equation}
      H^* =
      K^*_{\textbf{S}_0}
      C^*_{\textbf{P}_0}
      C^*_{\textbf{E}_0}
      C^*_{\textbf{U}_0}
      K^*_{\textbf{T}_0}
      K^*_{\textbf{S}_1}
      C^*_{\textbf{P}_1}
      C^*_{\textbf{E}_1}
      C^*_{\textbf{U}_1}
      K^*_{\textbf{T}_1}
      \dots
      K^*_{\textbf{S}_{d-1}}
      C^*_{\textbf{P}_{d-1}}
      C^*_{\textbf{E}_{d-1}}
      C^*_{\textbf{U}_{d-1}}
      K^*_{\textbf{T}_{d-1}}.
    \end{equation}
In practice, clearing the exchange buffers is implicit.  We illustrate the generalized, unbalanced forward and adjoint halo exchanges in Appendix~\ref{app:halo}.  This view justifies an observation that has been used in production PDE-constrained optimization codes for some time~\cite{ZhangHuangEtAl2013}: in the adjoint of halo exchange, there is an \textit{add} operation into the bulk tensor.  This is ultimately because the three copy operations, at the center of each part of the exchange, copy data from the bulk of one worker to the halo region of another and the ensuing adjoint phase must produce an add.

\paragraph{Implementation}

In our distributed deep learning library, DistDL, we have provided implementations of many necessary primitives for PyTorch \texttt{autograd}.  In parallel environments, verification of correctness using numerical gradient validation is difficult.  Fortunately, data movement operations are linear and we can exploit the fact that the forward operator is its own Jacobian, $F = \mathcal{F}$, and the definition of the adjoint to establish an equivalent test for correctness.  We say that an implementation of $F^*$ is coherent with $F$ if the adjoint test is satisfied,
\begin{align}
  \frac{\abs{\angles{F\vec{x}, \vec{y}}_{\CM^n} - \angles{\vec{x}, F^*\vec{y}}_{\CM^m}}}{\max\braces{\norm{F\vec{x}}_{\CM^n}\norm{\vec{y}}_{\CM^n},\norm{\vec{x}}_{\CM^m}\norm{F^*\vec{y}}_{\CM^m}}} < \varepsilon \;\;\forall \vec{x}\in\CM^m,\;\forall \vec{y}\in\CM^n.
\end{align}

\section{Model parallel layers}\label{sec:layers}

The parallel primitives defined in the previous section are sufficient for assembling implementations of common neural network layer functions.
We broadly categorize neural network layers into three classes: sparse layers, dense layers, and point-wise layers.  The classes are distinguished by the locality of interaction between degrees-of-freedom in the input tensor due to the layer function.  This locality determines which tensors are distributed, how they are distributed, and which parallel data movement primitives are necessary.  Point-wise layers, such as activation functions that operate on individual degrees-of-freedom, are embarrassingly parallel.  Native implementations of these functions can be used in distributed neural networks without further intervention and we omit them from the ensuing discussion.  While we give the algorithm for the adjoint pass of the composited distributed layer, we only have to provide the deep learning framework with the forward algorithm: all necessary adjoint data movement operations are already provided to the AD tool.
In the subsequent development, all rank-$d$ tensors are partitioned along each dimension by a $d$-length partition vector, which describes the number of workers in each dimension.

\paragraph{Sparse layers}

Sparse layers are characterized by the use of a small, sliding kernel function over the input tensor to map to output tensors.  Such functions include those with learnable network parameters, like the weights and biases in convolutional layers, and those without, such as pooling layers.

Among this class of layers, \emph{pooling} layers are the most straight-forward to parallelize.  Assume the input and output tensor $\vec{x}$ and $\vec{y}$ have feature-space dimension $D$ and over-all shape $n_b \times n_c \times m_0 \times \cdots \times m_{D-1}$ and $n_b \times n_c \times n_0 \times \cdots \times n_{D-1}$, where $n_b$, $n_c$, $m_i$, and $n_i$ are the batch, channel, and feature-space dimensions, respectively.  For both tensors, distributed over a partition $P$ with shape $1 \times P_c \times P_0 \times \cdots \times P_{D-1}$, the distributed pooling algorithm and the adjoint of its Jacobian are:

\hspace{5em}
\begin{minipage}[t]{.4\textwidth}
    Forward Pooling Algorithm
    \begin{algorithmic}[1]
      \State Input: $\vec{x}$ \textcolor{white}{$\delta$}
      \State $\vec{x} \gets H\vec{x}$  \textcolor{white}{$\delta$}
      \State $\vec{y} \gets \texttt{Pool(}\vec{x}\texttt{)}$  \textcolor{white}{$\delta$}
      \State Output: $\vec{y}$ \textcolor{white}{$\delta$}
    \end{algorithmic}
\end{minipage}%
\begin{minipage}[t]{.4\textwidth}
    Adjoint Pooling Algorithm
    \begin{algorithmic}[1]
      \State Input: $\delta \vec{y}$
      \State $\delta \vec{x} \gets [\delta\texttt{Pool}]^*\texttt{(}\delta \vec{y}\texttt{)}$
      \State $\delta\vec{x} \gets H^*\delta\vec{x}$
      \State Output: $\delta \vec{x}$
    \end{algorithmic}
\end{minipage}

The algorithm does not rely on linearity in the pooling operation, so any pooling operation is permitted, including average and max pooling.  The halo exchange $H$ is strongly dependent on the pooling kernel size, stride, dilation, and padding parameters.  In practice, padding and unpadding shims are required to address cases where halos are needed or extra input is provided (e.g., those in Appendix~\ref{app:halo}).


\emph{Convolutional} layers are a frequent target for parallelization~\cite{Shazeer2019} and were targeted by~\cite{DrydenMaruyamaEtAl2019a} to improve strong parallel scalability.  Ultimately, we seek weak scalability as we are interested in problems where the input tensors can have billions of degrees-of-freedom.  Anticipating that these tensors will be decomposed over potentially hundreds of workers, we avoid the explicit all-reduce operation often described.  Instead, we formulate the layer so that the all-reduce appears implicitly: a broadcast in the forward implementation naturally induces a sum-reduce in the adjoint phase.

Assume a similar structure as for the pooling layer, except that the learnable weights $\vec{w}$ have shape $n_{co} \times n_{ci} \times k_0 \times \cdots \times k_{D-1}$, where $n_{ci}$ and $n_{co}$ are the input and output channel sizes and $k_i$ is the kernel size, and are distributed over partition $P_r$ with shape $P_{co} \times P_{ci}$.
To avoid multiple counting of the bias, assume that the learnable part of the bias is only present on one $P_{co} \times 1$ subpartition of $P_r$.
For $P_x$ and $P_y$ with shapes $1 \times 1 \times P_{ci} \times P_0 \times \cdots \times P_{D-1}$ and $1 \times P_{co} \times 1 \times P_0 \times \cdots \times P_{D-1}$,\footnote{The additional dimensions aid the broadcasting pattern but do not impact the result.} a work partition $P_w$ with shape $1 \times P_{co} \times P_{ci} \times P_0 \times \cdots P_{D-1}$, and using broadcast and reduce operations that are similar to the NumPy broadcasting rules~\cite{Numpy2020}\footnote{The main difference is our broadcast is source-to-destination only, while NumPy broadcast is bi-directional.} along partitions, the generalized distributed convolution layer and the adjoint of its Jacobian are:

\hspace{5em}
\begin{minipage}[t]{.4\textwidth}
    Forward Convolution Algorithm
    \begin{algorithmic}[1]
      \State Input: $\vec{x}$ \textcolor{white}{$\delta$}
      \State $\vec{x} \gets H\vec{x}$  \textcolor{white}{$\delta$}
      \State $\vechat{w} \gets B_{\braces{P_r} \rightarrow \braces{P_w}}\vec{w}$  \textcolor{white}{$\delta$}
      \State $\vechat{b} \gets B_{\braces{P_r} \rightarrow \braces{P_w}}\vec{b}$  \textcolor{white}{$\delta$}
      \State $\vechat{x} \gets B_{\braces{P_x} \rightarrow \braces{P_w}}\vec{x}$  \textcolor{white}{$\delta$}
      \State $\vechat{y} \gets \texttt{Conv(}\vechat{w}, \vechat{b}; \vechat{x}\texttt{)}$  \textcolor{white}{$\delta$}
      \State $\vec{y} \gets R_{\braces{P_w} \rightarrow \braces{P_y}}\vechat{y}$  \textcolor{white}{$\delta$}
      \State Output: $\vec{y}$ \textcolor{white}{$\delta$}
    \end{algorithmic}
\end{minipage}%
\begin{minipage}[t]{.4\textwidth}
    Adjoint Convolution Algorithm
    \begin{algorithmic}[1]
      \State Input: $\delta \vec{y}$
      \State $\delta\vechat{y} \gets B_{\braces{P_y} \rightarrow \braces{P_w}}\delta\vec{y}$
      \State $\delta \vechat{w}, \delta \vechat{b}, \delta \vechat{x} \gets [\delta\texttt{Conv}]^*\texttt{(}\delta \vechat{y}\texttt{)}$
      \State $\delta \vec{x} \gets R_{\braces{P_w} \rightarrow \braces{P_x}}\delta \vechat{x}$
      \State $\delta \vec{b} \gets R_{\braces{P_w} \rightarrow \braces{P_r}}\delta \vechat{b}$
      \State $\delta \vec{w} \gets R_{\braces{P_w} \rightarrow \braces{P_r}}\delta \vechat{w}$
      \State $\delta\vec{x} \gets H^*\delta\vec{x}$
      \State Output: $\delta \vec{x}$
    \end{algorithmic}
\end{minipage}

If the tensors are distributed over the feature-space exclusively, or over channels exclusively, the algorithm can be significantly simplified by removing multiple broadcasts or reductions.  Distributed \emph{up-sampling} and \emph{down-sampling} layers are constructed similarly.

\paragraph{Dense layers}

Dense layers are characterized by full-connection between input and output degrees-of-freedom, often through the \emph{affine} function $\vec{y} = \vec{W}\vec{x} + \vec{b}$, where $\vec{W}$ is a dense $n_{fo} \times n_{fi}$ matrix and $n_{fo}$ and $n_{fi}$ are the number of output and input features.  Optimal parallelism in such layers is found through a distributed generalized matrix-matrix multiplication, or GEMM, algorithm~\cite{BosilcaGenetEtAl2018}.  Optimal GEMM structure and performance is dependent on the computing environment, the size and rank of the tensors, and is an area of open research.  We present an implementation based on the primitives above, recognizing that depending on the partitioning of workers, most production distributed GEMM implementations will have similar flavor.  A distributed affine layer has similar setup as the distributed convolution, except that the weight tensor is $n_{fo} \times n_{fi}$, where $n_{fo}$ and $n_{fi}$ are the number of features in and out, and is distributed on $P_w$, a $P_{fo} \times P_{fi}$ partition.  The learnable bias, of size $n_{fo}$, is present only on one $P_{fo} \times 1$ subset of $P_w$, to avoid any issue with multiple-counting of the bias.  For simplicity, we assume that the layer is fully-connected and that input and output tensors $\vec{x}$ and $\vec{y}$ have size $n_b \times n_{fi}$ and $n_b \times n_{fo}$, and are distributed on partitions $P_x$ and $P_y$, with shape $1 \times P_{fi}$ and $1 \times P_{fo}$, respectively.  The extension to arbitrary tensor dimensions is similar to  the distributed convolution layer.  The algorithm and the adjoint of its Jacobian are:

\hspace{5em}
\begin{minipage}[t]{.4\textwidth}
    Forward Affine Algorithm
    \begin{algorithmic}[1]
      \State Input: $\vec{x}$ \textcolor{white}{$\delta$}
      \State $\vechat{x} \gets B_{\braces{P_x} \rightarrow \braces{P_w}}\vec{x}$  \textcolor{white}{$\delta$}
      \State $\vechat{y} \gets \texttt{Affine(}\vechat{w}, \vechat{b}; \vechat{x}\texttt{)}$  \textcolor{white}{$\delta$}
      \State $\vec{y} \gets R_{\braces{P_w} \rightarrow \braces{P_y}}\vechat{y}$  \textcolor{white}{$\delta$}
      \State Output: $\vec{y}$ \textcolor{white}{$\delta$}
    \end{algorithmic}
\end{minipage}%
\begin{minipage}[t]{.4\textwidth}
    Adjoint Affine Algorithm
    \begin{algorithmic}[1]
      \State Input: $\delta \vec{y}$
      \State $\delta\vechat{y} \gets B_{\braces{P_y} \rightarrow \braces{P_w}}\delta\vec{y}$
      \State $\delta \vechat{w}, \delta \vechat{b}, \delta \vechat{x} \gets [\delta\texttt{Affine}]^*\texttt{(}\delta \vechat{y}\texttt{)}$
      \State $\delta \vec{x} \gets R_{\braces{P_w} \rightarrow \braces{P_x}}\delta \vechat{x}$
      \State Output: $\delta \vec{x}$
    \end{algorithmic}
\end{minipage}


\section{Example}\label{sec:examples}

\begin{figure}[h]
  \centering
  \includegraphics[height=1.3in,width=5.5in]{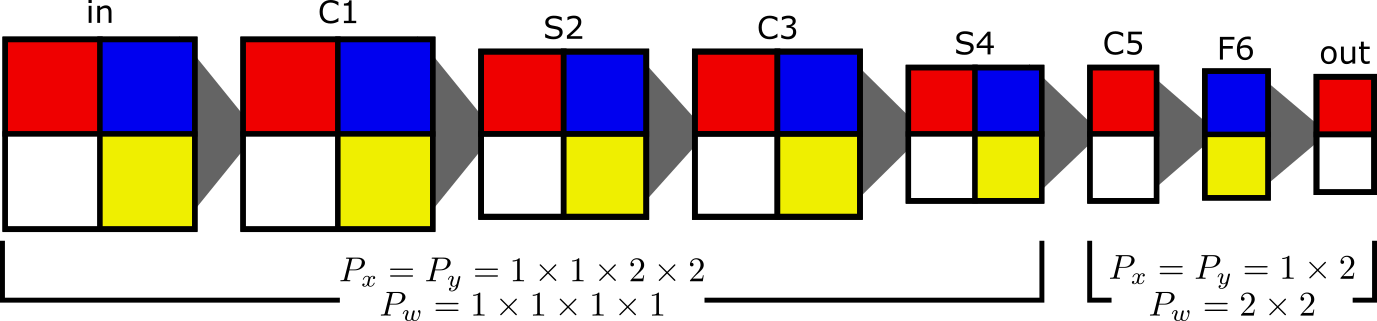}
  \caption{Global structure of a distributed Lenet-5 network in DistDL.}
  \label{fig:lenet-dist}
\end{figure}

We have implemented a number of the above layers in DistDL, as a demonstration of the effectiveness of our model.  These implementations explicitly rely on PyTorch's underlying implementation of the base layer function.  Using our distributed convolution, pooling, and affine layers, as well as some transpose layers as glue, we have implemented a distributed implementation of the Lenet-5 convolutional neural network~\cite{LecunBottouEtAl1998}.  We emphasize that this is not intended as a performance implementation, rather our aim was to validate the mathematical framework that we have developed.  The DNN itself is parallelized over a small number of workers (4), due to Lenet-5 and MNIST's tiny size, the underlying components satisfy adjoint tests for much larger tensors and  partitions.

Figure~\ref{fig:lenet-dist} shows a high-level diagram of the distributed network and worker-specific details are given in Appendix~\ref{app:network}, along with detailed experimental parameters.  Over 50 trials, training over the standard MNIST training data set with random initial network parameters, 10 epochs, and $n_b=256$, the sequential and distributed networks produce equivalent results: an average of $98.54\%$ and $98.55\%$ correct predictions on the standard test data set.

\section{Outlook \& Future Developments}\label{sec:outlook}

We have presented a linear-algebraic framework for\ data movement for distributed deep learning.  Using this framework, we have demonstrated that common neural network functions can be implemented in a distributed environment using the developed parallel primitives.  As a proof of concept, we have restricted our concrete implementation to a CPU implementation using MPI.
Concrete, performance driven realizations of these algorithms will be highly dependent on the target super computing architecture and the DNN structure.  For example, we anticipate that this framework will be particularly useful when applied to the large volumetric data sets and the physics-informed neural networks~\cite{YangPerdikaris2019,RaissiPerdikarisEtAl2019} currently being investigated for PDE constrained optimization, such as seismic inversion.
In any scenario, to achieve production-level performance on current extreme-scale, hybrid CPU-GPU supercomputers, and to leverage the increasing power of GPUs on smaller machines, a proper concrete implementation will need to be adapted to use Remote Direct Memory Access~\cite{LiuWuEtAl2004} (RDMA) or alternative interfaces to GPU-to-GPU communication technologies, such as the GPU support in recent MPI implementations~\cite{NVIDIA2017}.  This is an area of ongoing development.

\newpage
\section*{Broader Impact}

We anticipate that these developments will have no ethical or societal consequences distinct from any other development in high-performance computing (HPC) technology.  However, these developments provide a future path to democratize HPC technology with deep learning, the same way that the broad availability of PyTorch, Tensorflow, and cloud computing technologies have democratized ML.  Any positive or negative outcome arises strictly from the application selection of the user.  Only those with access to parallel computers will have immediate advantage from this work, though the model can be applied to local shared-memory computers, too.  If a parallel training job fails, lost time and money are the negative consequences.  There are no underlying biases present in this approach.

\begin{ack}
TJG was supported by the Luther and Alice Hamlett Undergraduate Research Support program.
\end{ack}


\bibliographystyle{ieeetr}
\bibliography{master}






\newpage
\begin{appendices}
\renewcommand\thefigure{\thesection\arabic{figure}}
\section{Derivations}\label{app:derivations}

\subsection{Derivation of the adjoint of allocation}

Assume the allocation $A: \CM^m \rightarrow \CM^n$, $\vec{x} = \begin{bmatrix}\vec{x}_a\end{bmatrix} \in \CM^m$, and $\vec{y} = \begin{bmatrix} \vec{y}_a \\ \vec{y}_b \end{bmatrix} \in \CM^n$.  Then, under the standard inner product, the adjoint of $A_b$ is,
\begin{align*}
\angles{A_b\vec{x},\vec{y}}_{\CM^n} &= \sum_{i=0}^{n-1}(A_b\vec{x})_iy_i = \sum_{i=0}^{m-1}(I_a\vec{x}_a)_iy_i = \sum_{i=0}^{m-1}x_iy_i \\
                                    &= \sum_{i=0}^{m-1}x_i(I_a\vec{y}_a)_i = \sum_{i=0}^{m-1}x_i(I_a\vec{y}_a + O_b^T\vec{y}_b)_i
                                    = \angles{\vec{x},A_b^T\vec{y}}_{\CM^m} = \angles{\vec{x},A_b^*\vec{y}}_{\CM^m}.
\end{align*}
Derivations of the adjoints of $K_b$ and $S_{a\rightarrow b}$ follow similarly for the standard inner product.

\subsection{Construction of copy}

The in-place copy operator, $C_{a \rightarrow b}: \CM^m \rightarrow \CM^m$, takes input $\vec{x} = \begin{bmatrix}\vec{x}_a \\ \vec{x}_b\end{bmatrix} \in \CM^m$ and produces output $\vec{x} = \begin{bmatrix}\vec{x}_a \\ \vec{x}_a\end{bmatrix} \in \CM^m$.  Thus,
\begin{align*}
  C_{a \rightarrow b} = \begin{bmatrix} I_a & O_b \\ I_a & O_b \end{bmatrix}
                      = \begin{bmatrix} I_a & O_b \\ I_a & I_b \end{bmatrix}\begin{bmatrix} I_a & O_b \\ O_a & O_b \end{bmatrix}
                      = S_{a \rightarrow b} K_b.
\end{align*}
Then, the adjoint is,
\begin{align*}
  C^*_{a \rightarrow b} = (S_{a \rightarrow b} K_b)^* =  K^*_bS^*_{a \rightarrow b} = K_bS_{b \rightarrow a}.
\end{align*}

The out-of-place copy operator, $C_{a \rightarrow b}: \CM^m \rightarrow \CM^n$, takes input $\vec{x} = \begin{bmatrix}\vec{x}_a \end{bmatrix} \in \CM^m$ and produces output $\vechat{x} = \begin{bmatrix}\vec{x}_a \\ \vec{x}_a\end{bmatrix} \in \CM^n$.  Thus,
\begin{align*}
  C_{a \rightarrow b} = \begin{bmatrix} I_a \\ I_a \end{bmatrix}
                      = \begin{bmatrix} I_a & O_b \\ I_a & I_b \end{bmatrix}\begin{bmatrix} I_a \\ O_b \end{bmatrix}
                      = S_{a \rightarrow b} A_b.
\end{align*}
Then, the adjoint is,
\begin{align*}
  C^*_{a \rightarrow b} = (S_{a \rightarrow b} A_b)^* =  A^*_bS^*_{a \rightarrow b} = D_bS_{b \rightarrow a}.
\end{align*}

\subsection{Construction of move}

The in-place move operator, $M_{a \rightarrow b}: \CM^m \rightarrow \CM^m$, takes input $\vec{x} = \begin{bmatrix}\vec{x}_a \\ \vec{x}_b\end{bmatrix} \in \CM^m$ and produces output $\vec{x} = \begin{bmatrix}\vec{0}_a \\ \vec{x}_a\end{bmatrix} \in \CM^m$.  Thus,
\begin{align*}
  M_{a \rightarrow b} = \begin{bmatrix} O_a & O_b \\ I_a & O_b \end{bmatrix}
                      = \begin{bmatrix} O_a & O_b \\ O_a & I_b \end{bmatrix}\begin{bmatrix} I_a & O_b \\ I_a & I_b \end{bmatrix}\begin{bmatrix} I_a & O_b \\ O_a & O_b \end{bmatrix}
                      = K_aS_{a \rightarrow b} K_b.
\end{align*}
Then, the adjoint is,
\begin{align*}
  M^*_{a \rightarrow b} = (K_a S_{a \rightarrow b} K_b)^* =  K^*_bS^*_{a \rightarrow b}K^*_a = K_bS_{b \rightarrow a}K_a.
\end{align*}

The out-of-place move operator, $M_{a \rightarrow b}: \CM^m \rightarrow \CM^{m^\prime}$, takes input $\vec{x} = \begin{bmatrix}\vec{x}_a \end{bmatrix} \in \CM^m$ and produces output $\vechat{x} = \begin{bmatrix}\vec{x}_a\end{bmatrix} \in \CM^{m^\prime}$, and $\CM^{m^\prime}$ is a different memory subset of the same size.  Thus,
\begin{align*}
  M_{a \rightarrow b} = \begin{bmatrix} O_a \\ I_a \end{bmatrix}
                      = \begin{bmatrix} O_a I_b \end{bmatrix} \begin{bmatrix} I_a & O_b \\ I_a & I_b \end{bmatrix}\begin{bmatrix} I_a \\ O_a \end{bmatrix}
                      = D_a S_{a \rightarrow b} A_b.
\end{align*}
Then, the adjoint is,
\begin{align*}
  M^*_{a \rightarrow b} = (D_a S_{a \rightarrow b} A_b)^* =  A^*_bS^*_{a \rightarrow b}D^*_a = D_bS_{b \rightarrow a}A_a.
\end{align*}

\section{Halo exchange}\label{app:halo}

\subsection{Irregularly structured halo regions}

The subsequent examples show the impact of different kernel parameters and tensor partitions on the halo regions for some different kernels and input sizes.  In each case, the driver for the computational load balance is the output distribution.  Consequently, absent any padding, assuming the input comes from another layer with the same property, the input is also balanced.  While we show 1-D examples for simplicity of presentation, the same patterns emerge in multidimensional cases, with more complex interactions between the halo regions.

In the following figures, bulk regions are illustrated in solid black lines and halo regions are given in dashed lines.  The numbers, arrows, and braces illustrate the access pattern.  Directional arrows indicate the input influence on output, numbers in the input tensor are indices, and numbers in the output tensor are the input index at the root of the kernel for that output index.
We have selected these examples for their representative behavior and the kernel parameters are commonly used in many DNNs.

\paragraph{``Normal'' convolution}

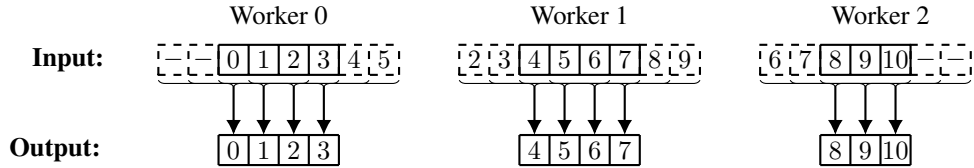
\begin{figure}[h]
  \centering
  \begin{tikzpicture}
		
	\node[] at (-2.0, 0.2) {\textbf{Input:}};
	\node[] at (-2.175, -1.0) {\textbf{Output:}};
	
	\begin{scope}[shift={(0, 0)}]
	
	\draw[step=0.4, black, thick] (0, 0) grid (1.6, 0.4);
	\draw[step=0.4, black, thick, dashed] (1.59, 0) grid (2.41, 0.41);
	\draw[step=0.4, black, thick, dashed] (-0.81, 0) grid (0, 0.41);
	
	\draw[step=0.4, black, thick] (0, -0.8) grid (1.6, -1.2);
	
	\foreach \startx / \stopx  in {-0.8/1.2, -0.4/1.6, 0/2, 0.4/2.4} {
		\draw[decoration={brace, mirror, raise=2pt}, decorate] (\startx, 0) -- (\stopx, 0);
	}
	
	\foreach \startx / \stopx in {0.2/0.2, 0.6/0.6, 1.0/1.0, 1.4/1.4} {
		\draw[-{Latex[width=2mm, length=2mm]}, thick] (\startx, -0.15) -- (\stopx, -0.8);
	}
	
	\foreach \x / \n in {-0.6/-, -0.2/-, 0.2/0, 0.6/1, 1.0/2, 1.4/3, 1.8/4, 2.2/5} {
		\node[minimum width=0.4] at (\x, 0.2) {$\n$};
	}
	
	\foreach \x / \n in {0.2/0, 0.6/1, 1.0/2, 1.4/3} {
		\node[minimum width=0.4] at (\x, -1.0) {$\n$};
	}

	\node[] at (0.8, 0.8) {Worker 0};
	
	\end{scope}
	
	\begin{scope}[shift={(4, 0)}]
	
	\draw[step=0.4, black, thick] (0, 0) grid (1.6, 0.4);
	\draw[step=0.4, black, thick, dashed] (1.59, 0) grid (2.41, 0.41);
	\draw[step=0.4, black, thick, dashed] (-0.81, 0) grid (0, 0.41);
	
	\draw[step=0.4, black, thick] (0, -0.8) grid (1.6, -1.2);
	
	\foreach \startx / \stopx  in {-0.8/1.2, -0.4/1.6, 0/2, 0.4/2.4} {
		\draw[decoration={brace, mirror, raise=2pt}, decorate] (\startx, 0) -- (\stopx, 0);
	}
	
	\foreach \startx / \stopx in {0.2/0.2, 0.6/0.6, 1.0/1.0, 1.4/1.4} {
		\draw[-{Latex[width=2mm, length=2mm]}, thick] (\startx, -0.15) -- (\stopx, -0.8);
	}
	
	\foreach \x / \n in {-0.6/2, -0.2/3, 0.2/4, 0.6/5, 1.0/6, 1.4/7, 1.8/8, 2.2/9} {
		\node[minimum width=0.4] at (\x, 0.2) {$\n$};
	}
	
	\foreach \x / \n in {0.2/4, 0.6/5, 1.0/6, 1.4/7} {
		\node[minimum width=0.4] at (\x, -1.0) {$\n$};
	}
	
	\node[] at (0.8, 0.8) {Worker 1};
	
	\end{scope}
	
	\begin{scope}[shift={(8, 0)}]
	
	\draw[step=0.4, black, thick] (0, 0) grid (1.2, 0.4);
	\draw[step=0.4, black, thick, dashed] (1.19, 0) grid (2.0, 0.41);
	\draw[step=0.4, black, thick, dashed] (-0.81, 0) grid (0, 0.41);
	
	\draw[step=0.4, black, thick] (0, -0.8) grid (1.2, -1.2);
	
	\foreach \startx / \stopx  in {-0.8/1.2, -0.4/1.6, 0/2} {
		\draw[decoration={brace, mirror, raise=2pt}, decorate] (\startx, 0) -- (\stopx, 0);
	}
	
	\foreach \startx / \stopx in {0.2/0.2, 0.6/0.6, 1.0/1.0} {
		\draw[-{Latex[width=2mm, length=2mm]}, thick] (\startx, -0.15) -- (\stopx, -0.8);
	}
	
	\foreach \x / \n in {-0.6/6, -0.2/7, 0.2/8, 0.6/9, 1.0/10, 1.4/-, 1.8/-} {
		\node[minimum width=0.4] at (\x, 0.2) {$\n$};
	}
	
	\foreach \x / \n in {0.2/8, 0.6/9, 1.0/10} {
		\node[minimum width=0.4] at (\x, -1.0) {$\n$};
	}
	
	\node[] at (0.8, 0.8) {Worker 2};
	
	\end{scope}

\end{tikzpicture}
  \caption{Uniform halo sizes induced by a $k=5$ centered kernel and width 2 padding.}
  \label{fig:halo-norm-conv}
\end{figure}

Assume a centered convolution kernel with size $k=5$, input tensor size $n=11$, partition size $P=3$, and assume a zero-padding of width 2 is implicitly added to the input boundaries.
In Figure~\ref{fig:halo-norm-conv} we illustrate that this situation yields the ``normal'', uniform halo sizes.

\paragraph{Unbalanced convolution}

\begin{figure}[h]
  \centering
  \begin{tikzpicture}
	
	\node[] at (-1.0, 0.2) {\textbf{Input:}};
	\node[] at (-1.175, -1.0) {\textbf{Output:}};

	\begin{scope}[shift={(0, 0)}]
	
	\draw[step=0.4, black, thick] (0, 0) grid (1.6, 0.4);
	\draw[step=0.4, black, thick, dashed] (1.59, 0) grid (2.81, 0.41);
	
	\draw[step=0.4, black, thick] (0.79, -0.8) grid (2.0, -1.2);
	
	\foreach \startx / \stopx  in {0/2, 0.4/2.4, 0.8/2.8} {
		\draw[decoration={brace, mirror, raise=2pt}, decorate] (\startx, 0) -- (\stopx, 0);
	}
	
	\foreach \startx / \stopx in {1.0/1.0, 1.4/1.4, 1.8/1.8} {
		\draw[-{Latex[width=2mm, length=2mm]}, thick] (\startx, -0.15) -- (\stopx, -0.8);
	}
	
	\foreach \x / \n in {0.2/0, 0.6/1, 1.0/2, 1.4/3, 1.8/4, 2.2/5, 2.6/6} {
		\node[minimum width=0.4] at (\x, 0.2) {$\n$};
	}
	
	\foreach \x / \n in {1.0/2, 1.4/3, 1.8/4} {
		\node[minimum width=0.4] at (\x, -1.0) {$\n$};
	}

	\node[] at (1.4, 0.8) {Worker 0};
	
	\end{scope}
	
	\begin{scope}[shift={(3.2, 0)}]
	
	\draw[step=0.4, black, thick, dashed] (-.01, 0) grid (0.4, 0.4);
	\draw[step=0.4, black, thick] (0.4, 0) grid (2.0, 0.4);
	\draw[step=0.4, black, thick, dashed] (2.0, 0) grid (2.41, 0.41);
	
	\draw[step=0.4, black, thick] (0.79, -0.8) grid (1.6, -1.2);
	
	\foreach \startx / \stopx in {0/2, 0.4/2.4} {
		\draw[decoration={brace, mirror, raise=2pt}, decorate] (\startx, 0) -- (\stopx, 0);
	}
	
	\foreach \startx / \stopx in {1.0/1.0, 1.4/1.4} {
		\draw[-{Latex[width=2mm, length=2mm]}, thick] (\startx, -0.15) -- (\stopx, -0.8);
	}
	
	\foreach \x / \n in {0.2/3, 0.6/4, 1.0/5, 1.4/6, 1.8/7, 2.2/8} {
		\node[minimum width=0.4] at (\x, 0.2) {$\n$};
	}
	
	\foreach \x / \n in {1.0/5, 1.4/6} {
		\node[minimum width=0.4] at (\x, -1.0) {$\n$};
	}

	\node[] at (1.2, 0.8) {Worker 1};
	
	\end{scope}
	
	\begin{scope}[shift={(6, 0)}]
	
	\draw[step=0.4, black, thick, dashed] (-.01, 0) grid (1.2, 0.4);
	\draw[step=0.4, black, thick] (1.2, 0) grid (2.4, 0.4);
	
	\draw[step=0.4, black, thick] (0.79, -0.8) grid (1.6, -1.2);
	
	\foreach \startx / \stopx in {0/2, 0.4/2.4} {
		\draw[decoration={brace, mirror, raise=2pt}, decorate] (\startx, 0) -- (\stopx, 0);
	}
	
	\foreach \startx / \stopx in {1.0/1.0, 1.4/1.4} {
		\draw[-{Latex[width=2mm, length=2mm]}, thick] (\startx, -0.15) -- (\stopx, -0.8);
	}
	
	\foreach \x / \n in {0.2/5, 0.6/6, 1.0/7, 1.4/8, 1.8/9, 2.2/10} {
		\node[minimum width=0.4] at (\x, 0.2) {$\n$};
	}
	
	\foreach \x / \n in {1.0/7, 1.4/8} {
		\node[minimum width=0.4] at (\x, -1.0) {$\n$};
	}

	\node[] at (1.2, 0.8) {Worker 2};
	
	\end{scope}
\end{tikzpicture}
  \caption{Non-uniform halo sizes induced by a $k=5$ centered kernel and no padding.}
  \label{fig:halo-unbalanced-conv}
\end{figure}
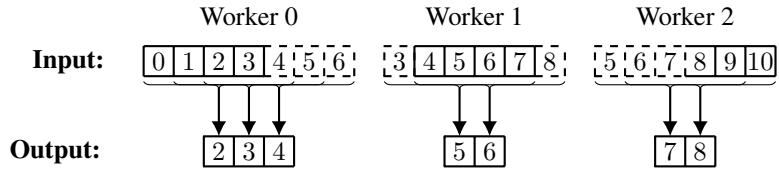

Assume a centered convolution kernel with size $k=5$, input tensor size $n=11$, partition size $P=3$, and assume a no padding is added to the input boundaries.  Then the output length is $m=7$.
In Figure~\ref{fig:halo-unbalanced-conv} we illustrate that this situation yields unbalanced halo sizes, where the first and last workers have large, one-sided halos and the middle worker has small, balanced halos.

\paragraph{Simple unbalanced pooling}

\begin{figure}[h]
  \centering
  \begin{tikzpicture}
	
	\node[] at (-1.0, 0.2) {\textbf{Input:}};
	\node[] at (-1.175, -1.0) {\textbf{Output:}};
	
	\begin{scope}[shift={(0, 0)}]
	
	\draw[step=0.4, black, thick] (0, 0) grid (1.6, 0.4);
	
	\draw[step=0.4, black, thick] (0.4, -0.8) grid (1.2, -1.2);
	
	\foreach \startx / \stopx  in {0.0/0.8, 0.8/1.6} {
		\draw[decoration={brace, mirror, raise=2pt}, decorate] (\startx, 0) -- (\stopx, 0);
	}
	
	\foreach \startx / \stopx in {0.4/0.6, 1.2/1.0} {
		\draw[-{Latex[width=2mm, length=2mm]}, thick] (\startx, -0.15) -- (\stopx, -0.8);
	}
	
	\foreach \x / \n in {0.2/0, 0.6/1, 1.0/2, 1.4/3} {
		\node[minimum width=0.4] at (\x, 0.2) {$\n$};
	}
	
	\foreach \x / \n in {0.6/0, 1.0/2} {
		\node[minimum width=0.4] at (\x, -1.0) {$\n$};
	}
	
	\node[] at (0.8, 0.8) {Worker 0};
	
	\end{scope}

	\begin{scope}[shift={(2.5, 0)}]
	
	\draw[step=0.4, black, thick] (0, 0) grid (1.2, 0.4);
	\draw[step=0.4, black, thick, dashed] (1.19, 0) grid (1.6, 0.41);
	
	\draw[step=0.4, black, thick] (0.4, -0.8) grid (1.2, -1.2);
	
	\foreach \startx / \stopx  in {0.0/0.8, 0.8/1.6} {
		\draw[decoration={brace, mirror, raise=2pt}, decorate] (\startx, 0) -- (\stopx, 0);
	}
	
	\foreach \startx / \stopx in {0.4/0.6, 1.2/1.0} {
		\draw[-{Latex[width=2mm, length=2mm]}, thick] (\startx, -0.15) -- (\stopx, -0.8);
	}
	
	\foreach \x / \n in {0.2/4, 0.6/5, 1.0/6, 1.4/7} {
		\node[minimum width=0.4] at (\x, 0.2) {$\n$};
	}
	
	\foreach \x / \n in {0.6/4, 1.0/6} {
		\node[minimum width=0.4] at (\x, -1.0) {$\n$};
	}
	
	\node[] at (0.8, 0.8) {Worker 1};
	
	\end{scope}

	\begin{scope}[shift={(4.6, 0)}]
	
	\draw[step=0.4, black, thick] (0.4, 0) grid (1.6, 0.4);
	
	\draw[step=0.4, black, thick] (0.79, -0.8) grid (1.2, -1.2);
	
	\foreach \startx / \stopx  in {0.8/1.6} {
		\draw[decoration={brace, mirror, raise=2pt}, decorate] (\startx, 0) -- (\stopx, 0);
	}
	
	\foreach \startx / \stopx in {1.2/1.0} {
		\draw[-{Latex[width=2mm, length=2mm]}, thick] (\startx, -0.15) -- (\stopx, -0.8);
	}
	
	\foreach \x / \n in {0.6/7, 1.0/8, 1.4/9} {
		\node[minimum width=0.4] at (\x, 0.2) {$\n$};
	}
	
	\foreach \x / \n in {1.0/8} {
		\node[minimum width=0.4] at (\x, -1.0) {$\n$};
	}
	
	\node[] at (1.0, 0.8) {Worker 2};
	
	\end{scope}

	\end{tikzpicture}
  \caption{Halo sizes induced by a $k=2$ right-looking kernel, with stride 2.}
  \label{fig:halo-pool-small}
\end{figure}
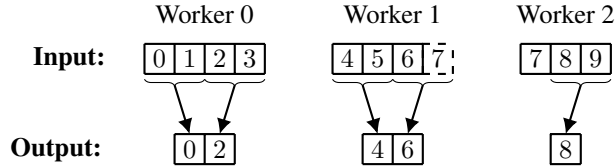

Assume a right-looking pooling kernel with size $k=2$, stride $s=2$, input tensor size $n=11$, partition size $P=3$, and  no padding or dilation.
In Figure~\ref{fig:halo-pool-small} we illustrate that this situation yields both unbalanced halos and unnecessary data in the input tensor.  For the first worker, there is no halo.  For the second worker, only the right-side has a halo, with size 1.  The last worker does not have any halo, but to produce the \textit{same} output as the sequential case for this input, the first entry of the input tensor actually has to be removed when the input is provided to the local pooling operator.

\paragraph{Complex unbalanced pooling}

\begin{figure}[h]
  \centering
    \begin{tikzpicture}
    
	\node[] at (-1.0, 0.2) {\textbf{Input:}};
	\node[] at (-1.175, -1.0) {\textbf{Output:}};
    
    \begin{scope}[shift={(0, 0)}]
        \draw[step=0.4, black, thick] (0, 0) grid (1.6, 0.4);
        \draw[step=0.4, black, thick] (0.39, -0.8) grid (1.2, -1.2);
        \foreach \x / \n in {0.2/0, 0.6/1, 1/2, 1.4/3} {
            \node[minimum width=0.4] at (\x, 0.2) {\small{$\n$}};
        }
        \foreach \x / \n in {0.6/0, 1/2} {
            \node[minimum width=0.4] at (\x, -1.0) {\small{$\n$}};
        }
        
        \draw[decoration={brace,mirror,raise=2pt},decorate, thick] (0.0, 0.0) -- (0.8, 0.0);
        \draw[decoration={brace,mirror,raise=2pt},decorate, thick] (0.8, 0.0) -- (1.6, 0.0);
        
        \draw[-{Latex[width=2mm, length=2mm]}, thick] (0.4, -0.15) -- (0.6, -0.8);
        \draw[-{Latex[width=2mm, length=2mm]}, thick] (1.2, -0.15) -- (1, -0.8);
        
        \node[] at (0.8, 0.7) {Worker 0};
    \end{scope}
    
    \begin{scope}[shift={(2, 0)}]
        \draw[step=0.4, black, thick] (0, 0) grid (1.6, 0.4);
        \draw[step=0.4, black, thick] (0.39, -0.8) grid (1.2, -1.2);
        \foreach \x / \n in {0.2/4, 0.6/5, 1/6, 1.4/7} {
            \node[minimum width=0.4] at (\x, 0.2) {\small{$\n$}};
        }
        \foreach \x / \n in {0.6/4, 1/6} {
            \node[minimum width=0.4] at (\x, -1.0) {\small{$\n$}};
        }
        
        \draw[decoration={brace,mirror,raise=2pt},decorate, thick] (0.0, 0.0) -- (0.8, 0.0);
        \draw[decoration={brace,mirror,raise=2pt},decorate, thick] (0.8, 0.0) -- (1.6, 0.0);
        
        \draw[-{Latex[width=2mm, length=2mm]}, thick] (0.4, -0.15) -- (0.6, -0.8);
        \draw[-{Latex[width=2mm, length=2mm]}, thick] (1.2, -0.15) -- (1, -0.8);
        
        \node[] at (0.8, 0.7) {Worker 1};
    \end{scope}
    
    \begin{scope}[shift={(4, 0)}]
        \draw[step=0.4, black, thick] (0, 0) grid (1.2, 0.4);
        \draw[step=0.4, black, thick, dashed] (1.2, 0.0) grid (1.6, 0.4);
        \draw[step=0.4, black, thick] (0.39, -0.8) grid (1.2, -1.2);
        \foreach \x / \n in {0.2/8, 0.6/9, 1/10, 1.4/11} {
            \node[minimum width=0.4] at (\x, 0.2) {\small{$\n$}};
        }
        \foreach \x / \n in {0.6/8, 1/10} {
            \node[minimum width=0.4] at (\x, -1.0) {\small{$\n$}};
        }
        
        \draw[decoration={brace,mirror,raise=2pt},decorate, thick] (0.0, 0.0) -- (0.8, 0.0);
        \draw[decoration={brace,mirror,raise=2pt},decorate, thick] (0.8, 0.0) -- (1.6, 0.0);
        
        \draw[-{Latex[width=2mm, length=2mm]}, thick] (0.4, -0.15) -- (0.6, -0.8);
        \draw[-{Latex[width=2mm, length=2mm]}, thick] (1.2, -0.15) -- (1, -0.8);
        
        \node[] at (0.8, 0.7) {Worker 2};
    \end{scope}
    
    \begin{scope}[shift={(6, 0)}]
        \draw[step=0.4, black, thick] (0, 0) grid (1.2, 0.4);
        \draw[step=0.4, black, thick, dashed] (1.2, 0.0) grid (2.0, 0.4);
        \draw[step=0.4, black, thick] (0.79, -0.8) grid (1.6, -1.2);
        \foreach \x / \n in {0.2/11, 0.6/12, 1/13, 1.4/14, 1.8/15} {
            \node[minimum width=0.4] at (\x, 0.2) {\small{$\n$}};
        }
        \foreach \x / \n in {1/12, 1.4/14} {
            \node[minimum width=0.4] at (\x, -1.0) {\small{$\n$}};
        }
        
        \draw[decoration={brace,mirror,raise=2pt},decorate, thick] (0.4, 0.0) -- (1.2, 0.0);
        \draw[decoration={brace,mirror,raise=2pt},decorate, thick] (1.2, 0.0) -- (2.0, 0.0);
        
        \draw[-{Latex[width=2mm, length=2mm]}, thick] (0.8, -0.15) -- (1, -0.8);
        \draw[-{Latex[width=2mm, length=2mm]}, thick] (1.6, -0.15) -- (1.4, -0.8);
        
        \node[] at (1.0, 0.7) {Worker 3};
    \end{scope}
    
    \begin{scope}[shift={(8.4, 0)}]
        \draw[step=0.4, black, thick] (0, 0) grid (1.2, 0.4);
        \draw[step=0.4, black, thick, dashed] (1.2, 0.0) grid (1.6, 0.4);
        \draw[step=0.4, black, thick] (0.79, -0.8) grid (1.2, -1.2);
        \foreach \x / \n in {0.2/14, 0.6/15, 1/16, 1.4/17} {
            \node[minimum width=0.4] at (\x, 0.2) {\small{$\n$}};
        }
        \foreach \x / \n in {1/16} {
            \node[minimum width=0.4] at (\x, -1.0) {\small{$\n$}};
        }
        
        \draw[decoration={brace,mirror,raise=2pt},decorate, thick] (0.8, 0.0) -- (1.6, 0.0);
 
        \draw[-{Latex[width=2mm, length=2mm]}, thick] (1.2, -0.15) -- (1, -0.8);
        
        \node[] at (0.8, 0.7) {Worker 4};
    \end{scope}
    
    \begin{scope}[shift={(10.4, 0)}]
        \draw[step=0.4, black, thick] (0, 0) grid (1.2, 0.4);
        \draw[step=0.4, black, thick] (0.39, -0.8) grid (0.8, -1.2);
        \foreach \x / \n in {0.2/17, 0.6/18, 1/19} {
            \node[minimum width=0.4] at (\x, 0.2) {\small{$\n$}};
        }
        \foreach \x / \n in {0.6/18} {
            \node[minimum width=0.4] at (\x, -1.0) {\small{$\n$}};
        }
        
        \draw[decoration={brace,mirror,raise=2pt},decorate, thick] (0.4, 0.0) -- (1.2, 0.0);
        
        \draw[-{Latex[width=2mm, length=2mm]}, thick] (0.8, -0.15) -- (0.6, -0.8);
        
        \node[] at (0.8, 0.7) {Worker 5};
    \end{scope}
    
\end{tikzpicture}
  \caption{Halo sizes induced by a $k=2$ right-looking kernel, with stride 2.}
  \label{fig:halo-pool-large}
\end{figure}
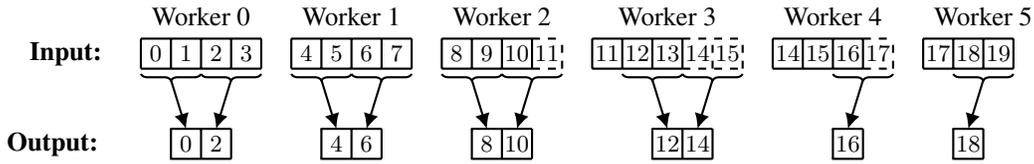

Assume a right-looking pooling kernel with size $k=2$, stride $s=2$, input tensor size $n=20$, partition size $P=6$, and no padding or dilation.
In Figure~\ref{fig:halo-pool-large} we illustrate that this situation yields many ranks with unbalanced halos and unnecessary data in the input tensor.  For the first and second workers, there are no halos.  The third worker has a right halo but no left halo.
The 4\textsuperscript{th} worker has 1 extra input on the left and a halo of length 2 on the right.
The 5\textsuperscript{th} worker has 2 extra input on the left and a halo of length 1 on the right.
The final worker has no halos, but one extra input on the left.  In cases with extra input data, those entries of the input tensor actually has to be removed when the input is provided to the local pooling operator.

\subsection{Generalized tensor halo exchange}

\begin{figure}[h]
  \centering
  \includegraphics[scale=0.8]{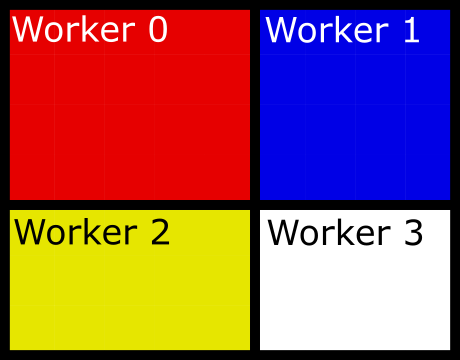}
  \caption{Data before forward halo exchange for $P=2 \times 2 $ partition of a rank-2 tensor.}
  \label{fig:halo-ex-start}
\end{figure}

\begin{figure}[h]
  \centering
  \begin{tabular}{cc}
  \begin{subfigure}[b]{0.47\textwidth}
    \includegraphics[scale=0.8]{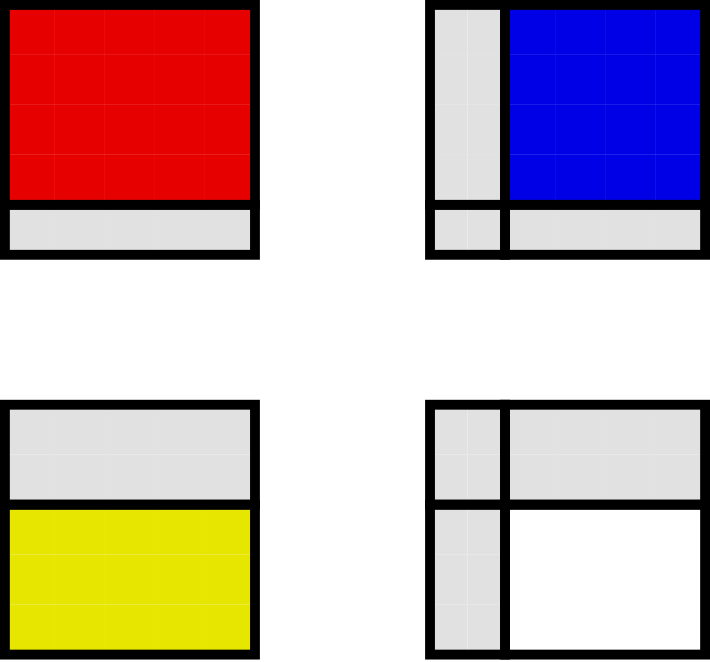}
    \caption{Setup of forward halo exchange.}
    \label{fig:halo-ex-fwd-a}
  \end{subfigure} &
  \begin{subfigure}[b]{0.47\textwidth}
    \includegraphics[scale=0.8]{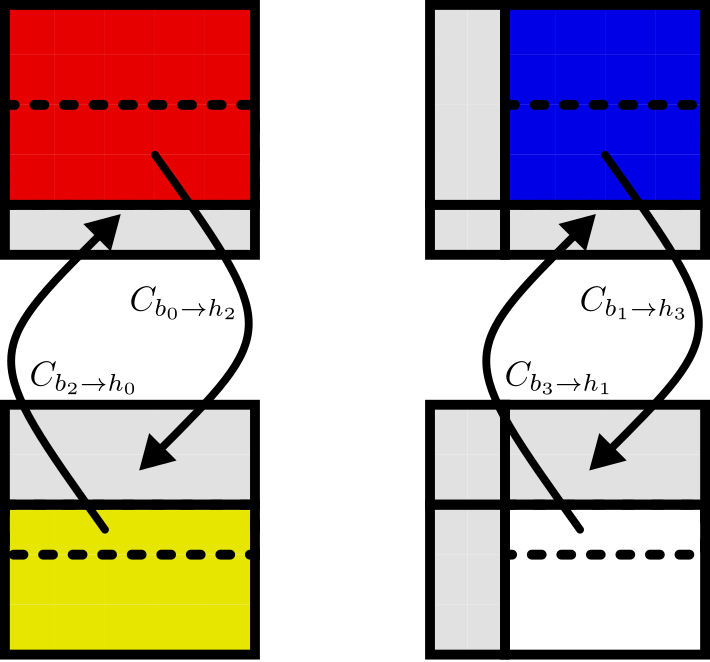}
    \caption{First phase of halo exchange.}
    \label{fig:halo-ex-fwd-b}
  \end{subfigure} \\&\\
  \begin{subfigure}[b]{0.47\textwidth}
    \includegraphics[scale=0.8]{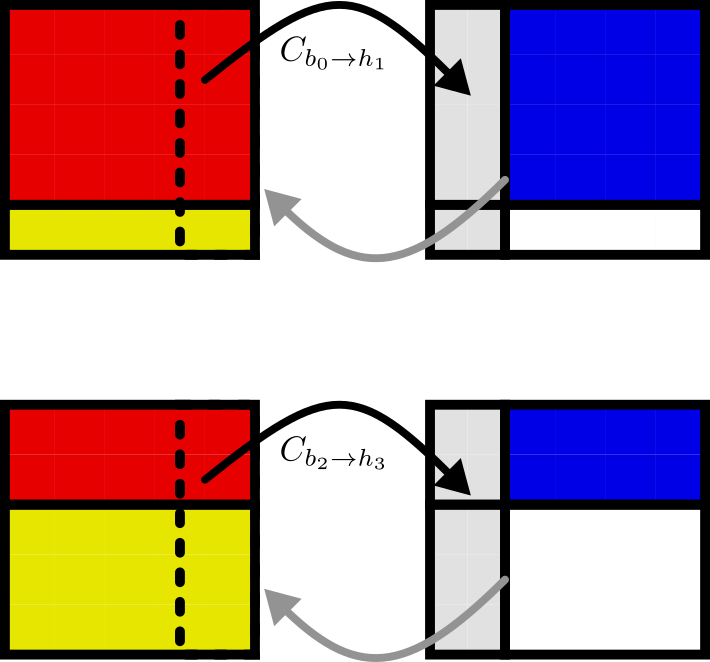}
    \caption{Second phase of halo exchange.}
    \label{fig:halo-ex-fwd-c}
  \end{subfigure} &
  \begin{subfigure}[b]{0.47\textwidth}
    \includegraphics[scale=0.8]{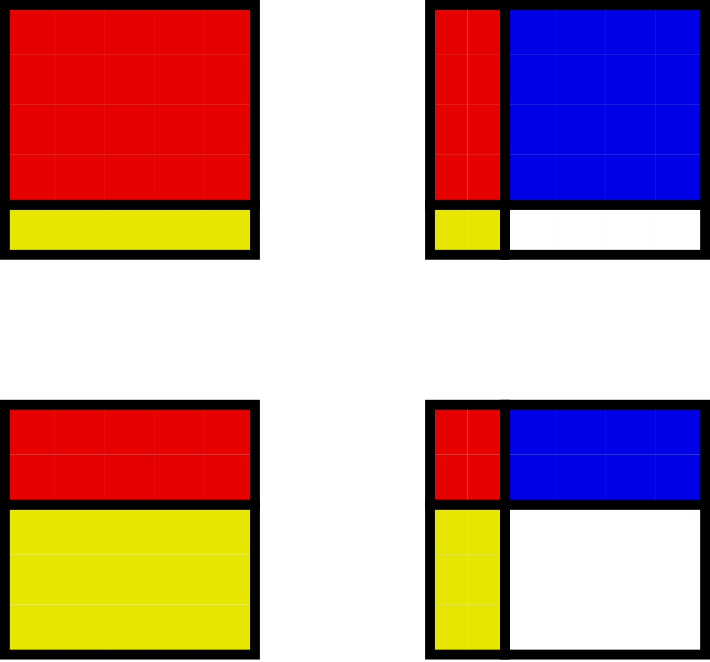}
    \caption{Result of forward halo exchange.}
    \label{fig:halo-ex-fwd-d}
  \end{subfigure}
  \end{tabular}
  \caption{Forward unbalanced halo exchange for $P=2 \times 2 $ partition of a rank-2 tensor.}
  \label{fig:halo-ex-fwd}
\end{figure}

Here we illustrate the generalized, unbalanced halo exchange on a rank-$2$ tensor, partitioned by a $P = 2 \times 2$ partition.  While the algorithm works for tensors of arbitrary rank with arbitrary partitions, a rank-$2$ tensor is sufficient to illustrate the concept.  In Figure~\ref{fig:halo-ex-start}, we have partitioned the tensor into 4 unequal, but load-balanced domains.  The colors will be maintained throughout subsequent figures to help illustrate data ownership.  The differences in size are exaggerated for clarity.  As seen in Figure~\ref{fig:halo-ex-fwd-a}, where gray regions are the halo region, workers 0 and 2 require no data from workers 1 and 3, but share width 3 data with them, workers 0 and 1 require width 2 data from workers 2 and 3, workers 2 and 3 require width 4 data from workers 0 and 1, and there are interior halos only.  We have chosen the vertical dimension to perform the first exchange.

Figure~\ref{fig:halo-ex-fwd} illustrates the sequence of copy operations in the forward halo exchange algorithm.  After two steps (Figures~\ref{fig:halo-ex-fwd-b} and~\ref{fig:halo-ex-fwd-c}, the final exchanged result is in Figure~\ref{fig:halo-ex-fwd-d}.  The exchange pattern is nested to minimize communication volume, as for larger, higher-rank tensors these volumes grow quickly.  The gray arrows in the second exchange phase indicate that no data needs to be shared.  We have omitted the action on the send and receive buffers, for clarity.

\begin{figure}[h]
  \centering
  \begin{tabular}{cc}
  \begin{subfigure}[b]{0.47\textwidth}
    \includegraphics[scale=0.8]{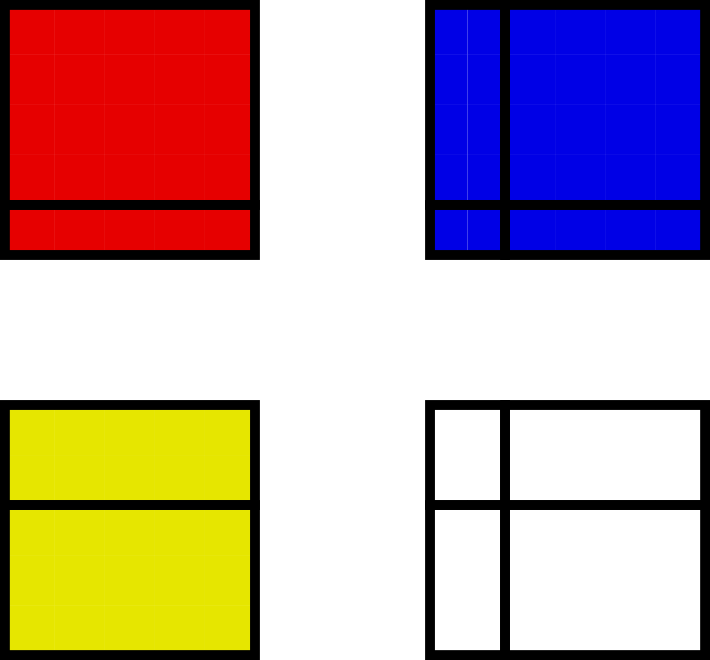}
    \caption{Setup of adjoint halo exchange.}
    \label{fig:halo-ex-adj-a}
  \end{subfigure} &
  \begin{subfigure}[b]{0.47\textwidth}
    \includegraphics[scale=0.8]{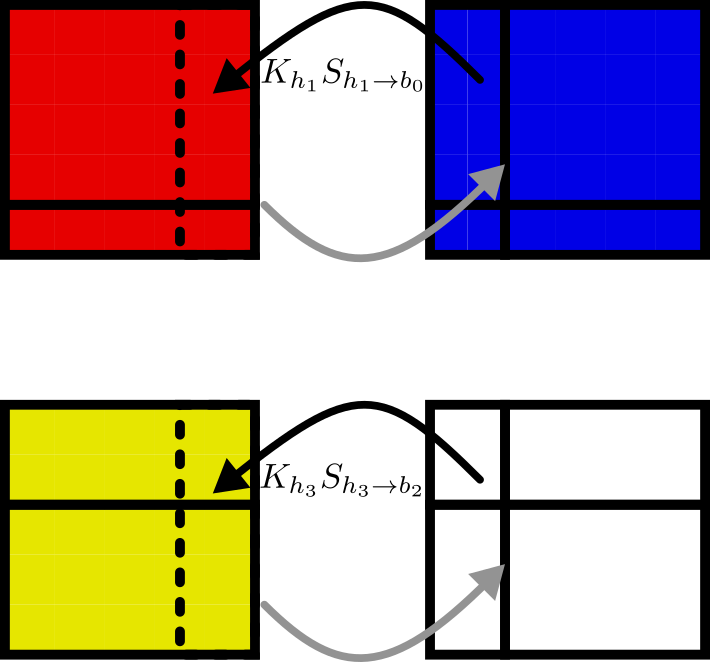}
    \caption{Adjoint of second phase halo exchange.}
    \label{fig:halo-ex-adj-b}
  \end{subfigure} \\&\\
  \begin{subfigure}[b]{0.47\textwidth}
    \includegraphics[scale=0.8]{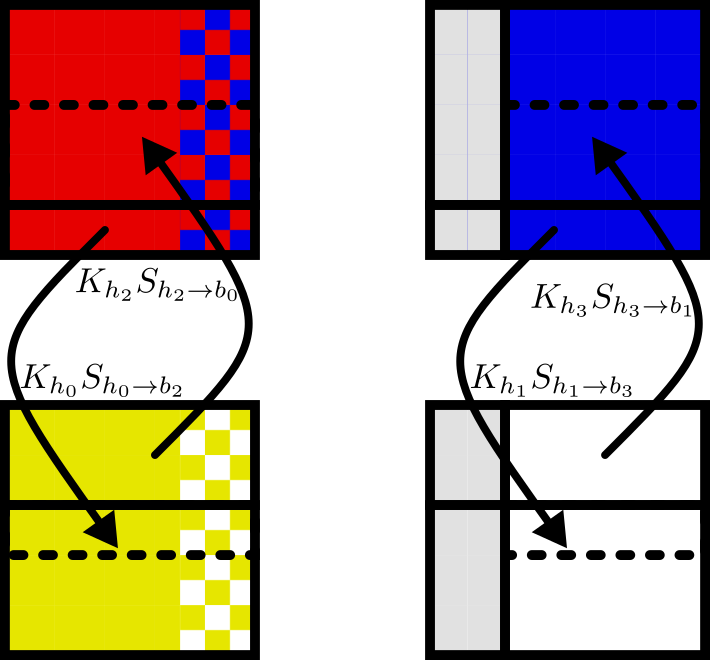}
    \caption{Adjoint of first phase halo exchange.}
    \label{fig:halo-ex-adj-c}
  \end{subfigure} &
  \begin{subfigure}[b]{0.47\textwidth}
    \includegraphics[scale=0.8]{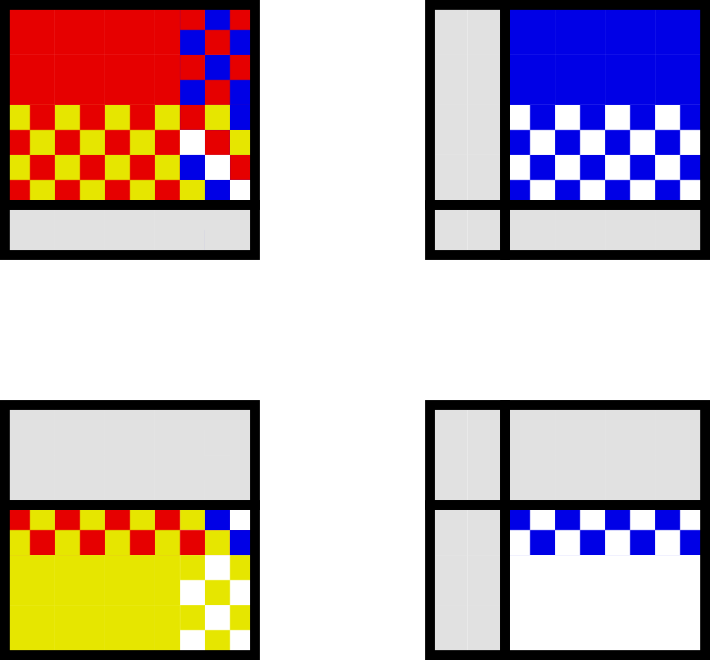}
    \caption{Result of adjoint halo exchange.}
    \label{fig:halo-ex-adj-d}
  \end{subfigure}
  \end{tabular}
  \caption{Adjoint unbalanced halo exchange for $P=2 \times 2 $ partition of a rank-2 tensor.}
  \label{fig:halo-ex-adj}
\end{figure}

Figure~\ref{fig:halo-ex-asj} illustrates the sequence of add-clear operations in the adjoint halo exchange algorithm.  Figure~\ref{fig:halo-ex-adj-a} shows the starting state, where each rank has input data starting in its halo regions.  After two steps (Figures~\ref{fig:halo-ex-asj-b} and~\ref{fig:halo-ex-asj-c}, the final exchanged result is in Figure~\ref{fig:halo-ex-asj-d}.  The checkerboard patterns indicate summation.  The gray arrows in the adjoint of the second exchange phase indicate that no data needs to be shared.  We have omitted the action on the send and receive buffers, for clarity.

\begin{figure}[h]
  \centering
  \includegraphics[scale=0.8]{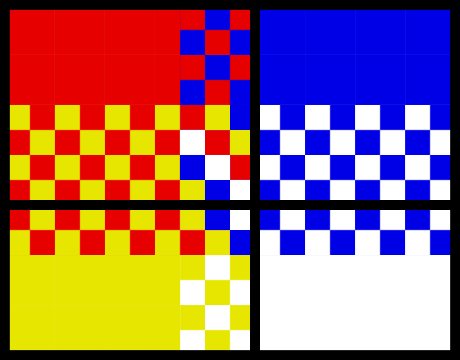}
  \caption{Data after adjoint halo exchange for $P=2 \times 2 $ partition of a rank-2 tensor.}
  \label{fig:halo-ex-end}
\end{figure}

\section{Distributed Lenet-5}\label{app:network}

\subsection{Global network architecture}

\begin{figure}[h]
  \centering
  \includegraphics[]{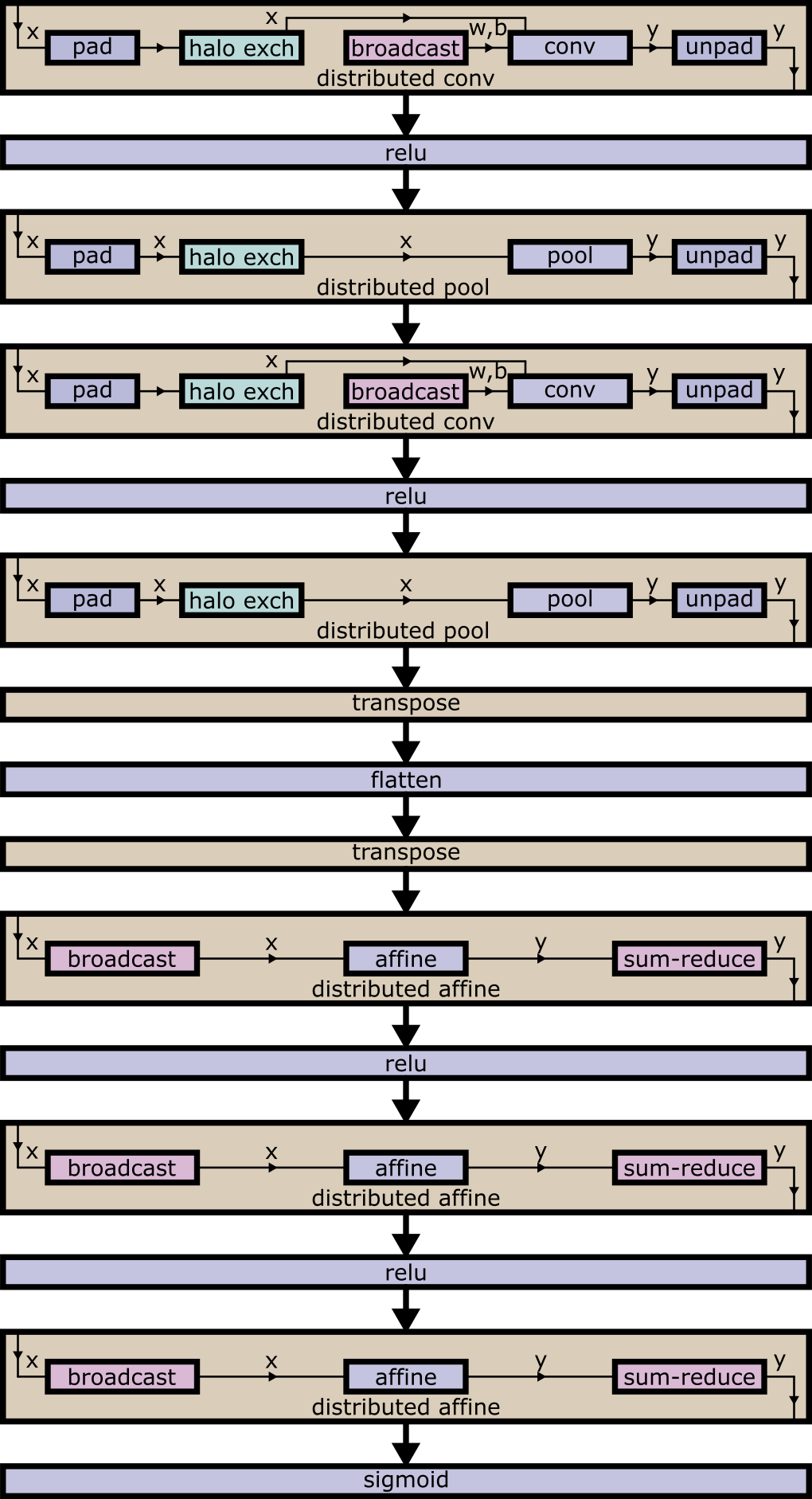}
  \caption{Global view of the distributed Lenet-5.  Sequential layers and those from the underlying deep learning framework are purple, distributed layers and primitives are brown, green, and pink.}
  \label{fig:distlenetfull}
\end{figure}

In Figure~\ref{fig:distlenetfull} we show the full structure of the distributed Lenet-5 network, including all necessary shims and sub-layers.  The parallel distribution of learnable parameters is provided in Table~\ref{tab:paramcount}.  The transpose layers are used to create better load balance on the inputs and outputs and their selection is system and implementation dependent.  We also make use of transpose layers to distribute input data and collect outputs (not shown).

\begin{table}[h]
  \centering
  \begin{tabular}{rrllll}
     Layer  &Function & Worker 0        & Worker 1     & Worker 2     & Worker 3 \\
    C1      &Conv     & $w: (6,1,5,5)$  & None         & None         & None \\
            &         & $b: (6)$        &              &              &      \\
    S2      &Pool     & None            & None         & None         & None \\
    C3      &Conv     & $w: (16,6,5,5)$ & None         & None         & None \\
            &         & $b: (16)$       &              &              &      \\
    S4      &Pool     & None            & None         & None         & None \\
    C5      &Affine & $w: (60,200)$     & $w: (60,200)$ & $w: (60,200)$ & $w: (60,200)$ \\
            &       & $b: (60)$         &               & $b: (60)$     &               \\
    F6      &Affine & $w: (42,60)$      & $w: (42,60)$  & $w: (42,60)$  & $w: (42,60)$ \\
            &       & $b: (42)$         &               & $b: (42)$     &              \\
    Output  &Affine & $w: (5,42)$       & $w: (5,42)$   & $w: (5,42)$   & $w: (5,42)$ \\
            &       & $b: (5)$          &               & $b:(5)$       &
  \end{tabular}

\caption{Learnable parameters per worker, per layer.}
\label{tab:paramcount}
\end{table}

\subsection{Experimental parameters}

The experimental parameters for comparing sequential and distributed versions of Lenet-5 are reported here.
The training data consists of the standard 60,000 MNIST training data set, broken into batches of size 256.  Because the distributed network requires a fixed batch size, the final 96 images are dropped from the data set, for both networks.
The test data consists of the standard 10,000 MNIST test data set, broken into batches of size 256.  Because the distributed network requires a fixed batch size, the final 96 images are dropped from the data set, for both networks.
The sequential network is implemented in PyTorch using PyTorch's native neural network modules.  The distributed network is implemented using DistDL's distributed neural network models, as shown in Figure~\ref{fig:distlenetfull}.  Each network was trained 50 times, with random initial weights, over 10 epochs.  The Adam optimizer, with learning rate $\alpha=0.001$, was applied to the cross-entropy loss function.  This experiment was run on an Intel Xeon E3-1505M with 32 GB of RAM.  Source implementation has been tested and verified on Virginia Tech's Cascades cluster.

\end{appendices}
\end{document}